\documentclass[]{interact}
\usepackage{siunitx}
\usepackage{amsmath,amssymb,amsfonts} 
\usepackage{times} 
\usepackage{tabularx}
\usepackage{threeparttable}
\usepackage{subfigure}
\usepackage[square, comma, sort&compress, numbers]{natbib}
\usepackage[pagebackref=false,breaklinks=true,letterpaper=true,colorlinks,bookmarks=false]{hyperref}
\usepackage{times}
\usepackage{breakurl}
\usepackage{array}
\usepackage{verbatim}
\usepackage{algorithm}
\usepackage{bbding}
\usepackage{algpseudocode}
\usepackage{lettrine}

\usepackage{bm} 
\usepackage{epstopdf}
\usepackage{subfigure}
\usepackage{multirow}
\usepackage{natbib}
\bibpunct[, ]{(}{)}{;}{a}{}{,}
\renewcommand\bibfont{\fontsize{10}{12}\selectfont}

\theoremstyle{plain}

\theoremstyle{definition}

\theoremstyle{remark}

\begin{document}
\title{Spatio-Temporal Bi-directional Cross-frame Memory for Distractor Filtering Point Cloud Single Object Tracking}
\author{
    \name{Shaoyu Sun\textsuperscript{a}\thanks{CONTACT Shaoyu Sun. Email: christyuws@sina.com} Chunyang Wang\textsuperscript{a,b,*}\thanks{*Corresponding author: Chunyang Wang, Email: wangchunyang19@163.com} Xuelian Liu\textsuperscript{b}\thanks{CONTACT Xuelian Liu.Email: tearlxl@126.com} Chunhao Shi\textsuperscript{c} \thanks{CONTACT Chunhao Shi. Email:shichunhaocust@163.com} Yueyang Ding\textsuperscript{a}\thanks{CONTACT Yueyang Ding.Email:15585666767@163.com}and Guan Xi\textsuperscript{b} \thanks{CONTACT Guan Xi. Email:15939168068@163.com }}
    \affil{\textsuperscript{a}School of Electronic and Information Engineering, Changchun University of Science and Technology, Changchun, China; \textsuperscript{b}Xi’an Key Laboratory of Active Photoelectric Imaging Detection Technology, Xi'an Technological University, Xi'an, China; \textsuperscript{c}Hong Kong Applied Science and Technology Research Institute, Hong Kong, China}}
    
\maketitle  

\begin{abstract}
3D single object tracking within LIDAR point clouds is a pivotal task in computer vision, with profound implications for autonomous driving and robotics. 
However, existing methods, which depend solely on appearance matching via Siamese networks or utilize motion information from successive frames, encounter significant challenges. Issues such as similar objects nearby or occlusions can result in tracker drift.
To mitigate these challenges, we design an innovative spatio-temporal bi-directional cross-frame distractor filtering tracker, named STMD-Tracker. Our first step involves the creation of a 4D multi-frame spatio-temporal graph convolution backbone. This design separates KNN graph spatial embedding and incorporates 1D temporal convolution, effectively capturing temporal fluctuations and spatio-temporal information.
Subsequently, we devise a novel bi-directional cross-frame memory procedure. This integrates future and synthetic past frame memory to enhance the current memory, thereby improving the accuracy of iteration-based tracking. This iterative memory update mechanism allows our tracker to dynamically compensate for information in the current frame, effectively reducing tracker drift.
Lastly, we construct spatially reliable Gaussian masks on the fused features to eliminate distractor points. This is further supplemented by an object-aware sampling strategy, which bolsters the efficiency and precision of object localization, thereby reducing tracking errors caused by distractors.
Our extensive experiments on KITTI, NuScenes and Waymo datasets demonstrate that our approach significantly surpasses the current state-of-the-art methods.
\end{abstract}


\section{Introduction}
Single object tracking in point clouds~\citep{MTM-Tracker,Feng23,MLSET,wang23,STNet} is vital in domains such as autonomous driving, robotics, and augmented reality. While deep learning approaches have significantly advanced 2D  single object tracking, translating these advancements to 3D point cloud single object tracking remains a challenging task, particularly due to the sparsity of point cloud data. Despite notable progress in this area, as evidenced by recent studies, tracking performance is still heavily impacted by occlusion or similar objects, which causes significant tracker drift.

\begin{figure}[t]
\centering
\includegraphics[width=14cm, height=7.2cm]{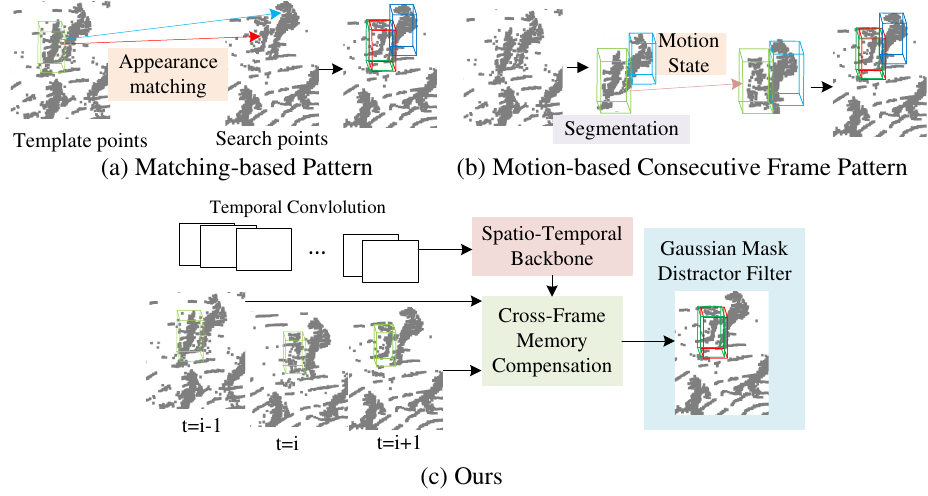}
\vspace{-0.15in}
\caption{\textbf{Comparison of 3D single object tracker pattern.} (a) Matching-based tracker utilize appearance matching to locate target. (b) Motion-based method use segmentation to capture motion pattern to predict the subsequent movement. (c) Our approach multi-frame spatio-temporal backbone with temporal convolution and cross-frame memory compensate for lost target information, utilizing Gaussian mask to filter out distractor for accurate localization.}
\label{fig:introduciton_fig}
\vspace{-0.1in}
\end{figure} 

 As shown in Figure.~\ref{fig:introduciton_fig} (a), previous Siamese trackers primarily using siamese-based matching method and recognition method~\citep{emrn, hsgm} to investigate the integration of target features, including P2B~\citep{P2B}, PTTR~\citep{PTTR}, LTTR~\citep{LTTR}, GLT-T~\citep{GLT-T} methods.  For example,  TAT~\citep{TAT} introduces a temporal-aware siamese tracker that leverages historical frame templates for feature propagation yet limited performance by overlooking latest frame contextual information. M2Track~\citep{M2Track} focuses on adjacent frames' motion information, yet its need for motion transformation and lack of localization context limits its effectiveness (in Figure.~\ref{fig:introduciton_fig}(b)). CXTrack~\citep{CXTrack} enhances tracking by using spatial contextual information and a local center embedding localization network. These methods~\citep{P2B,tcrl,LTTR,CXTrack} only transfer target cues from the latest frame to the current one, thus ignoring the abundant information in other historical frames. Moreover, They overlook the rich temporal semantic information in historical frames and sensitive to appearance variation caused by occlusion or distractor.

 Consequently, researches utilize multi-frames to propagate target cues but most are using motion information of consecutive frames, as shown in Figure.~\ref{fig:introduciton_fig} (b). STTracker~\citep{STTracker} leverages multi-frame feature encoding, but its localization relies on a BEV-based representation, which leads to significant information loss for small targets, leading to potential localization inaccuracies. CAT~\citep{Gao23} adaptively generates search and templates region based on point cloud density and employs masks for localization, however, direct augmentation with historical frame encoding may bring background noise. StreamTrack~\citep{StreamTrack} leverage cross-frame memory bank and a transformer to model long-range relationship for feature extraction, but its reliance on contrastive learning for similarity may inadequately filter out similar distractor. MBPTrack~\citep{MBPTrack} use multi-frame historical frames and continuously creates memory from last frame to interact with the current frame. However, most are using motion information of consecutive frames, which not directly build relationship of spatio-temporal and may still exit tracker drift due to occlusion or similar distractor.

As previously discussed, our objective is to capture the dynamic temporal changes of objects, as illustrated in Figure.~\ref{fig:introduciton_fig} (c). To this end, we have developed a novel tracking system, the STMD-Tracker. This system introduces multi-frame temporal encoding by integrating 1D temporal convolution into the 3D Backbone. This integration enriches the spatio-temporal encoding features by establishing relationships between spatial and temporal dimensions across multiple frames.
To tackle the issue of tracker drift, which is primarily caused by accumulated tracking errors, we have designed a bi-directional cross-frame memory module. This module utilizes both future frames and synthetic past frame memory to generate cross-frame memory, effectively eliminating distractor tracker drift. Upon integration with the current frame, this memory generates a synthetic current frame memory, which then serves as the past frame memory in subsequent iterations.
We have also implemented a Gaussian mask to filter out distractor points. By calculating the Gaussian distribution of the current frame's point coordinates and the previous frame's predicted bounding box coordinates, we can filter out points that are distant from the target. This mask is then applied as a weight on the features output from the previous stage's memory, resulting in a spatial semantic representation with more targeted attributes.
Lastly, we have designed an effective sampling method based on objectness score ranking. This method generates high-quality proposals, enhancing the accuracy of the predicted target center and the precision of bounding box localization.

In summary, our contributions are as follows:

\begin{itemize}
\item We propose a novel tracking system, the STMD-Tracker, which introduces multi-frame temporal encoding by integrating 1D temporal convolution into the 3D Backbone. This enriches the spatio-temporal encoding features by establishing relationships between spatial and temporal dimensions across multiple frames.
\item We present a bi-directional cross-frame memory module to tackle the issue of tracker drift caused by accumulated tracking errors. This module utilizes both future frames and synthetic past frame memory to generate cross-frame memory, effectively eliminating distractor tracker drift. 
\item We implement a Gaussian mask to filter out distractor points, resulting in a spatial semantic representation with more targeted attributes. This mask is applied as a weight on the features output from the previous stage's memory.
\item We design an effective sampling method based on objectness score ranking. This method generates high-quality proposals, enhancing the accuracy of the predicted target center and the precision of bounding box localization.
\end{itemize}

\section{related work}

\subsection{3D Single Object Tracking} 
In the field of point cloud single-object tracking (SOT), the common approach is based on Siamese networks like PTT~\citep{PTT}, STNet~\citep{STNet}, GLT-T~\citep{GLT-T} and GLT-T++ ~\citep{GLT-T++} use search and template region do feature matching for accurate tracking. Breaking away from the feature matching approach of Siamese networks, M2Track~\citep{M2Track} uses two consecutive frames motion information to track object based on segmentation, significantly improve the tracking capabilities. TAT~\citep{TAT} introduced a temporal-aware siamese single object tracking algorithm that selects high-quality target templates from historical frames for reliable propagation of specific target features, yet limited performance due to neglect the object around contextual information of the latest frame. STTracker~\citep{STTracker} employs multi-frame inputs separately fed into the backbone network, followed by a similarity-based spatio-temporal fusion module to track object. StreamTrack~\citep{StreamTrack} argues that two frames overlook long-term continuous motion information, hence it uses multiple continuous frames as a memory bank with a hybrid attention mechanism, while incorporating a contrastive sequence enhancement strategy to improve discrimination against false positives. 
CFIL \citep{weng2023cross} proposes a frequency-domain feature extraction module and feature interaction in the frequency domain to enhance salient features. MFC \citep{qiao2022novel} proposes a frequency-domain filtering module to achieve dense target feature enhancement. MTM-Tracker~\citep{MTM-Tracker} combine feature matching and motion modeling, by design iterative processment it not only capture short-term and also long-short motion for tracking.  MBPTrack~\citep{MBPTrack} also use multi-frames and memory iteration mechanism to tracking object. However, due to the similarity objects exit or occlusion by distractor result in tracker drift. This will lead error from historical frames iteration of feature representation. Our method, addressing this issue by introduce future frame and last frame to compensate current lost object information due to tracker drift. We also add 1D temporal convolution subequently by 3D spatial feature extraciton to total catpure temporal varaiation at the same location.

\subsection{Distractor Filtering in Object Tracking}
 For multi-object video tracking, DMtrack~\citep{DMTrack} employs dynamic convolution for lightweight one-shot detection of targets, distinguishing targets from distractor through re-identification ~\citep{pbsl} and inferring distractor by maintaining trajectories of potential similarity. For example, HPGN~\citep{hpgn} proposes a novel pyramid graph network targeting features, which is closely connected behind the backbone network to explore multi-scale spatial structural features. GiT~\citep{git}  proposes a structure where graphs and transformers interact constantly, enabling close collaboration between global and local features. Unlike the approach of using appearance models to filter out distractor, LaSOT~\citep{LaSOT} tracks distractor frame by frame, differentiating them from targets instead of ignoring them. DADTracker~\citep{DADTracker} leverages a relational attention mechanism~\citep{li2022enhancing} to aggregate features based on appearance relationships, effectively handling occlusions and appearance variations for improved tracking accuracy. In the single-object tracking domain, SiamCDA~\citep{SiamCDA} utilizes siamese-based foreground-background classification to differentiate the tracking target from distractor. SiamRN~\citep{SiamRN} employs meta-learning to filter out background distractor and introduces a contrastive training strategy to learn not only to match the same target but also to distinguish different objects. StreamTrack~\citep{StreamTrack} also use contrastive learning to do distractor-aware tracking. OPSNet~\citep{OPSNet} addressing the issue of distractor-induced mis-localization, combining a loss function and an object-preserved sampling strategy to improve RPN performance, with center mapping prediction and location error punishment, enhancing the accuracy of center localization. 
 In our method, we replace focus on distractor-aware tracking for distractor filter tracking with objectness score ranking sampling for high quality proposal. 
\section{Methods}

\begin{figure*}[t]
\centering
\includegraphics[width=1\linewidth]{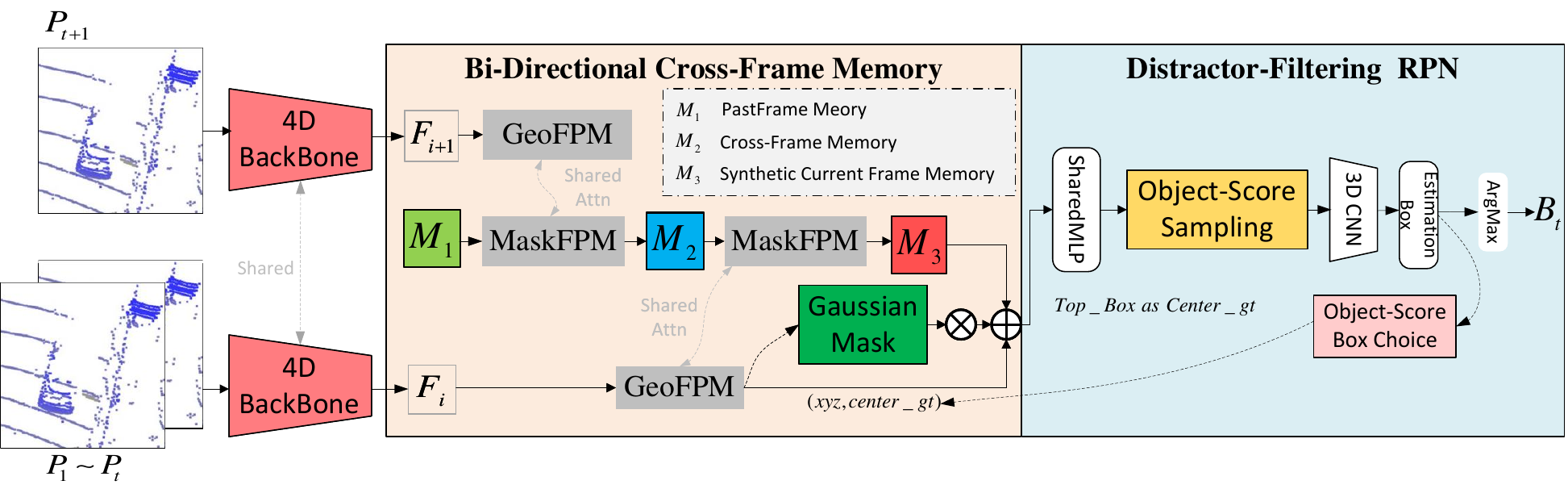}
 \vspace{-0.15in}
\caption{\textbf{The overall architecture of STMD-Tracker.} 
It comprises three main blocks, processing N frames through a \textit{4D spatio-temporal graph convolution backbone}. Initially, we utilize a 3D graph convolution backbone to embed spatial features. Subsequently, we apply 1D temporal convolution to embed multi-frame spatial features. These features are then fed into a \textit{ bi-directional cross-frame memory module}, which compensates for missing information in the current frame, thereby mitigating tracker drift. Finally, we employ a \textit{distractor filtering RPN} to eliminate false negative proposals, enhancing the accuracy of the tracking process.}
\label{fig:architecture}
 \vspace{-0.1in}
\end{figure*}

\subsection{overview}
In the domain of 3D object tracking, the bounding box of a tracking object is characterized by its central coordinates \((x, y, z)\), dimensions (width \(w\), length \(l\), height \(h\)), and orientation angle $\theta$. The objective in tracking scenarios is to track the  object across successive point cloud frames. Given the object's dimensions are established in the initial frame, the focus shifts to predict the central coordinates and orientation angle $\theta$. Temporal dynamics play a pivotal role in tracking accuracy and distinguishing between similar objects. As depicted in Figure.~\ref{fig:architecture}, we propose a spatio-temporal cross-frame memory distractor filtering tracker. Initially, a sequence of N raw point cloud frames is fed into a 4D spatial-temporal graph convolution backbone, which performs spatial neighborhood convolutions followed by a 1D temporal convolution along the time dimension. The bi-directional cross-frame memory module, designed to leverage the rich information from historical frames and not merely consider two consecutive frames, employs a dual updating mechanism with future and past frames to compensate for the occlusions of tracking objects by distractor in the current frame. To further localize the object, a distractor filtering Region Proposal Network (RPN) is employed to eliminate false negative box candidates, thus ensuring the precise localization of the object. The following sections will examine the complexities of each framework component in detail.

\subsection{4D Multi-Frames Spatio-Temporal Backbone}
In point cloud single object tracking, addressing temporal variations is a crucial challenge, as the appearance and position of the object can undergo significant changes over time. These variations become increasingly complex in sparse and incomplete point cloud data acquired from LiDAR, particularly for distant targets with minimal points. To effectively capture temporal variation, we introduce temporal convolution. Inspired by PSTNet~\citep{PSTNet}, which involves decoupling spatial and temporal convolutions due to their orthogonal and independent properties. This separation facilitates the alignment of features across various timestamps, significantly improving the network’s proficiency in capturing the spatio-temporal features of the object. Building on this, we have engineered a 4D multi-frame spatio-temporal graph convolution backbone to overcome these point cloud-related complexities.

\textbf{4D Backbone Formulation.}
Let $P_t \in \mathbb{R}^{3 \times N}$ and $F_t \in \mathbb{R}^{C \times N}$ denote point coordinates and features of the $t$-th frame in point cloud sequences, where N and C denote the number of points and feature channels, respectively. The 4D backbone convolution is formulated as:
\begin{equation}
F'_{t}(x, y, z) = \sum_{k=-\left\lfloor \frac{L}{2} \right\rfloor}^{\left\lfloor \frac{L}{2} \right\rfloor} 
\sum_{i=1}^{N} \Big( W_{k}(\delta_{x_i}, \delta_{y_i}, \delta_{z_i})
\cdot F_{t+k}(x+\delta_{x_i}, y+\delta_{y_i}, z+\delta_{z_i}) \Big), 
\end{equation}
where $(\delta_{x_i}, \delta_{y_i}, \delta_{z_i})$ represents the displacement of the $i$-th nearest neighbor point relative to the center point $(x, y, z)$. Moreover, because space and time are orthogonal and independent of each other, we further decompose spatial and temporal modeling as:
\begin{align}
F'_{t}(x, y, z) = \sum_{k=-\left\lfloor \frac{L}{2} \right\rfloor}^{\left\lfloor \frac{L}{2} \right\rfloor} 
\sum_{i=1}^{N} \Big( T_{k}(\delta_{x_i}, \delta_{y_i}, \delta_{z_i}) \nonumber \cdot \quad S_{k}(\delta_{x_i}, \delta_{y_i},\delta_{z_i})\\ 
 \cdot F_{t+k}(x+\delta_{x_i}, y+\delta_{y_i}, z+\delta_{z_i}) \Big).
\end{align}
The convolution kernel $W \in \mathbb{R}^{C' \times C \times l} \times R$ is decomposed into a spatial convolution kernel $S \in \mathbb{R}^{C_m \times C \times l} \times R$ and a temporal convolution kernel $T \in \mathbb{R}^{C' \times C_m \times l} \times R$ and $C_m$ is the dimension of the intermediate feature.
Given a point cloud sequence, we first use set abstraction layer\citep{PoinetNet++} to embed new center features and coordinates, and then we construct K nearest neighbour (KNN) graph to capture edge geometric features of local spatial structures in each frame. 

\textbf{3D Saptial Feature Embedding Formulation.} In the context of our model, the embedded geometric features after being processed by the EdgeConv\citep{EdgeConv} are denoted by $M^{'}_{t}$, which is a four-dimensional tensor with dimensions corresponding to 
\begin{align}
M'_{t}(x,y,z) = \sum_{i=1}^{N} \Big( &S(\delta_{x_i}, \delta_{y_i}, \delta_{z_i}) \cdot F_{t}(x+\delta_{x_i},y+\delta_{y_i},z+\delta_{z_i}) \Big),
\end{align}
where $ M^{'}_{t}(x,y,z) \in \mathbb{R}^{b \times l \times f \times p}$. Here $b$ is the batch size that represents the number of point cloud sequences processed simultaneously; $l$ refers to the number of frames and indicates the temporal extent of the data; $f$ is the number of feature channels,encapsulating the dimension of the extracted features; and $p$ signifies the number of points per frame, which defines the spatial resolution of the point cloud. In our spatial feature embedding part, $(\delta_{x_i}, \delta_{y_i}, \delta_{z_i})$ signifies the displacement from a point $(x, y, z)$ to its $i$-th nearest neighbor in a point cloud. This differs from traditional convolution in Euclidean domains-such as image processing, where displacement involves filter traversal over a feature map. Here, displacement captures relative position differences for edge feature construction in the graph-based processing of point clouds, which inherently possess non-Euclidean geometry.

\begin{figure}[t]
    \centering
    \includegraphics[width=0.65\linewidth]{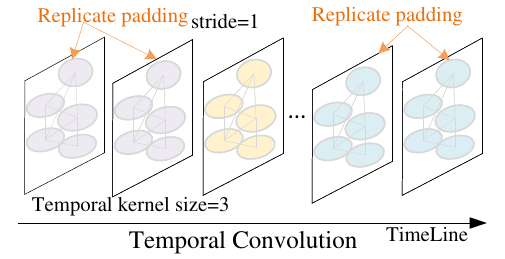}
    \vspace{-0.05in}
    \caption{\textbf{Temporal Convolution on Multi-Frame Spatial Feature Maps.} Each frame constructs a KNN graph to aggregate neighboring features for representing node characteristics. After embedding the spatial features of each frame, a 1D temporal convolution is applied along the temporal dimension. Purple and blue denote the replication of the first and last frames, respectively, serving as padding for the convolution. With a kernel size of 3 and stride of 1, the convolution of 8 frames captures both short-term motion and long-term tracking information.}
    \vspace{-0.1in}
    \label{fig:temporal convolution}
\end{figure}
\textbf{1D Temporal Convolution on Frames.}
After embedding the spatial features of each point in the cloud, as shown in Figure.\ref{fig:temporal convolution} we process the sequence of point clouds using temporal convolution. This aims to describe the local temporal dynamics of the point cloud sequences along the time dimension.
\begin{align}
F^{'}_{t}(x,y,z) = \sum_{k=-r_{t}}^{r_{t}} T_{k} \cdot (M^{'}_{t}(x,y,z)).
\end{align}

$F^{'}_{t}(x,y,z)$ means 1D temporal convolution feature embedding, $T_{k}$ means temporal kernel. To ensure temporal continuity and accommodate the edge frames within our 4D spatio-temporal model, we apply replicate padding to the left side of the first frame and the right side of the eighth frame. Subsequently, we conduct temporal convolution over the frames with a kernel size of 3 and a stride of 1. After passing through a Multi-layer Perceptron (MLP) for encoding, we ultimately obtain an 8 frames feature representation denoted as $F^{'}_{t}(x,y,z)\in \mathbb{R}^{b \times l \times f \times p}$. The size of the spatio-temporal feature embedding is the same as that of the spatial embedding. By integrating analyses across spatial and temporal dimensions, our approach accurately tracks target objects, even in dynamic and challenging environments. The incorporation of temporal convolutions not only enhances the understanding of the movement and appearance changes of the target object but also provides a comprehensive perspective for precise tracking in dynamic scenarios. 

\subsection{Bi-Directional Cross-Frame Memory Module}

\begin{figure}[t]
    \centering
    \includegraphics[width=1\linewidth]{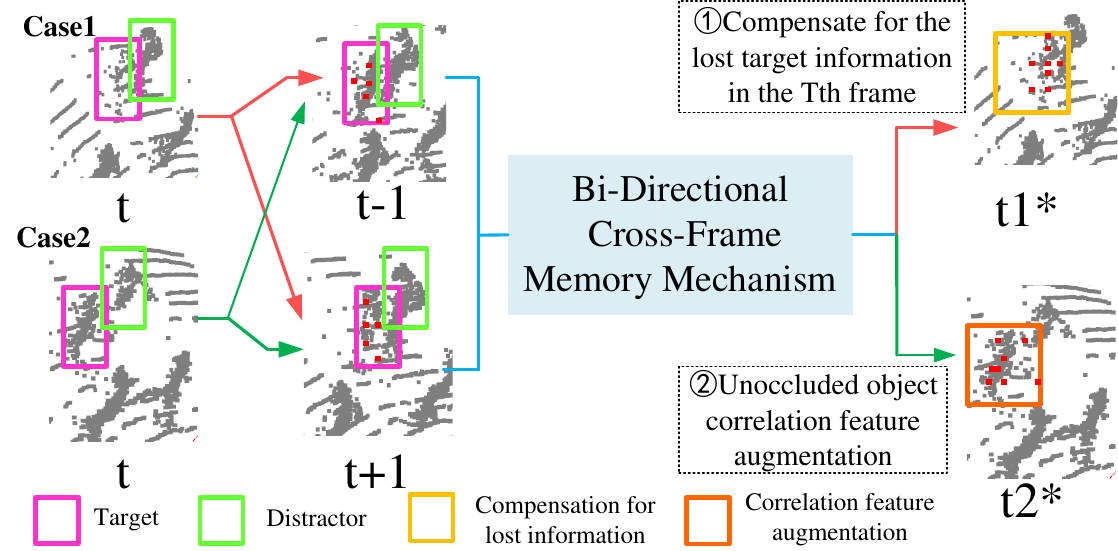}
    \vspace{-0.15in}
    \caption{\textbf{Bi-directional cross-frame memory module}. In Case 1, where the target in frame t has lost information due to occlusion. By inputting the subsequent frame (t+1) and the previous frame (t-1) into the bi-directional cross-frame memory module, which can compensate target information. In Case 2, when the target in frame t is unoccluded, the same bi-directional memory mechanism takes  frames t+1 and t-1 to augment the target's correlation feature.  Our method can eliminate tracker drift and improve tracking performance.}
    \vspace{-0.1in}
    \label{fig:memory}
\end{figure}
Due to the interference from similar objects and the sparsity of object points, the tracking object may disappear in a certain frame. While, both future and historic frame have richer semantic information which can help the tracker drift to distractor, the object can re-localized to compensate for its disappearance in the current frame. Therefore, as shown in Figure.~\ref{fig:memory}, we introduce a bi-directional cross-frame memory module into the 3D SOT task, which utilizes the future frame and synthetic past memory frame to generate cross-frame memory in reverse. Then, it works with the current frame to generate current memory in a forward manner. This current memory will be used as synthetic past memory in the next iteration.

\textbf{First Frame Past Memory Generation.} After processing through the Backbone, we obtain an 8 frames spatio-temporal feature embedding. During the propagation inference at time step $i< n_{\text{sam\_frame}} - 1$, For\(i=0\), $n_{\text{sam\_frame}} = 8$, $i = 0$, the first frame duplicates itself to serve as the future frame. In Equation~\eqref{eq:Transfomer features}, these 8 frames features then separately input into the GEO-FPM~\citep{MBPTrack} transformer layer to propagate to obtain a cross-frame transformer feature, defined as ${T_(0,0)}$, P means propagate operation. The process of $T_{(0,0)}$ is defined as follow:
\begin{equation}
 T_{(0,0)} = P(F_0, F_0).
\label{eq:Transfomer features}
\end{equation} 
Subsequently, the 0 frame transformer feature is input into the Mask-FPM~\citep{MBPTrack} transformer layer with first frame $F_0$ to obtain a mask and generate cross-frame memory, which is defined as ${M_(0,0)}$. In Equation~\eqref{eq:cross-frame memory} U means update operation, M means memory of the update result. The process of $M_{(0,0)}$ is defined as follow:
\begin{equation}
 M_{(0,0)} = U(T_{(0,0)}, F_0). 
\label{eq:cross-frame memory}
\end{equation}

\textbf{Bi-Directional Cross-Frame Memory Module.} When  $i< n_{\text{sam\_frame}} - 1$, for the $i^{th}$ memory, after the generation of synthetic past frame memory from the first frame, in Equation~\eqref{eq:future frame} it is aggregated with the future frame $F_{(i+1)}$ input into GeoFPM~\citep{MBPTrack} to propagate transformer features $T_{(i_1,i+1)}$. The process defined as below:
\begin{equation}
 T_{(i+1,i-1)} = P(F_{(i+1)}, M_{(i-1,i-1)}).
\label{eq:future frame}
\end{equation}

In Equation~\eqref{eq:ith cross-frame memory}, then we fed cross-frame transformer feature $T_{(i_1,i+1)}$, $i+1^{th}$ frame and synthetic current frame memory $M_{(i-1,i-1)}$ into MaskFPM~\citep{MBPTrack} to obtain mask features, which is cross-frame memory and the operation is update. This process updates the memory to cross-frame memory, represented as $M_{(i-1,i-1)}$. The process defined as fellow:
\begin{equation}
M_{(i+,i-1)} = U(T_{(i+1,i-1)}, F_{i+1}, M_{(i-1,i-1)}).
\label{eq:ith cross-frame memory}
\end{equation}

This cross-frame memory is used for forward updating of the current frame. In Equation~\eqref{eq:forward sythentic current frame memory} utilizing this cross-frame memory $ M_{(i+1,i-1)}$, along with the current frame $F_{(i)}$ input into GeoFPM, propagating a synthetic current frame transformer feature, indicated as $T_{(i,i')}$. 
\begin{equation}
 T_{(i,i')} = P(F_{(i)}, M_{(i+1,i-1)}). 
\label{eq:forward sythentic current frame memory}
\end{equation}

Then, in Equation~\eqref{eq:backward current memory} by inputting the synthetic current frame transformer feature $T_{(i,i')}$, cross-frame memory $M_{(i-1,i-1)}$, and the current frame $F_{(i)}$ into MaskFPM, we obtain mask features and thus update the memory features to synthetic current frame memory, denoted as $M_{i,i'}$. This operation analogous to a backward update of the current frame. And this synthetic current frame memory serves as the synthetic past frame memory for the next iteration. Therefore, this module is termed a bi-directional cross-frame memory generation module. The process defined as below:
\begin{equation}
 M_{(i,i')} = U(T_{(i,i')}, F_{(i)}, M_{(i-1,i-1)}). 
\label{eq:backward current memory}
\end{equation}

\textbf{Last Frame Feature Propagation.}
When \(i = 7 \), it does not satisfy the condition \( i < n_{\text{sam\_frame}}\), in Equation~\eqref{last frame} inputting the current frame \( F_i\) and the memory \( M_{(i-1,i-1)} \) into GeoFPM to obtain the transformer feature \(T_{(i,i-1)}\), which does not participate in the update of the current frame memory, so the update operation is skipped. The iteration is finished. The expression for the operation is as follows:
\begin{equation}
 T_{(i,i-1)} = P(F_{(i)}, M_{(i-1,i-1)}). 
 \label{last frame}
\end{equation}

By designing a bi-directional cross-frame memory module that continuously updates the memory with information from future and past frames, we can jointly compensate the object points in the current frame that are missing due to occlusion by similar objects. Or, in unoccluded condition, augment the target with correlation features. It effectively eliminates tracker drift and improve tracking performance.

\subsection{Gaussian Spatial Reliable Localization Network}
\begin{figure}[t]
    \centering    
\includegraphics[width=0.8\linewidth]{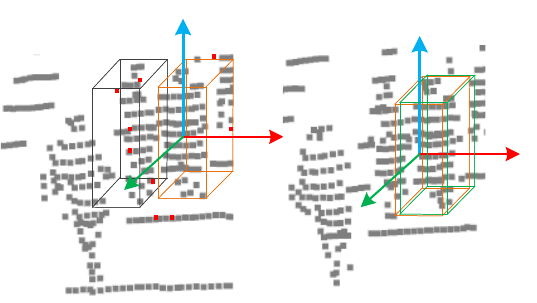}
 \vspace{-0.05in}
    \caption{\textbf{Gaussian mask filter distractor point.} The red points, representing distractor points, can be filtered out using a Gaussian mask. The incorrect prediction is marked with blank box, and correct prediction are marked with green box.}
    \vspace{-0.1in}
    \label{fig:gaussain mask}
\end{figure}

To further address the issue of tracker drift caused by occlusion of similar object near the tracked target, we design a Gaussian mask localization network (GMLocNet) in Figure.~\ref{fig:gaussain mask}. It effectively generate a Gaussian mask for each point by employing Gaussian function to assess points' distribution within spatial domain. Subsequently, it assigns mask as weights on features of last stage memory according to their spatial distribution. Furthermore, we also design an object-aware sampling to refine the selection of object points to generate high quality proposals. GMLocNet strategically emphasizes points close to the object's center meanwhile diminishing the influence of target irrelevant distractor points, significantly enhancing tracking performance. 

\textbf{Gaussian Mask Calculation.} In Equation~\eqref{gaussian mask calculation.}, The initial step involves computing the Euclidean distance between the coordinates of points in the current frame and the predicted bounding box coordinates of the last frame. The process of calculation can be formulated as:
\begin{equation}
G_\text{m} = e^{-\frac{(x-y)^2}{2\sigma^2}},
\label{gaussian mask calculation.}
\end{equation}

where $x\in \mathbb{R}^{3}$,$y\in \mathbb{R}^{3}$ represent the spatial coordinates of the current object point and previous estimation box, respectively. 
For the initial frame, the ground truth object coordinate ${y}_1= C_{\text{in}}$ is utilized to calculate the Gaussian mask.
For subsequent frames, the center coordinates of the box of the highest predicted objectness score $BBox_{\text{top}}$ is used as y (previous estimation box), which can be defined as ${y}_{t-1} = \text{BBox}_{\text{top}}$.

$(x-y)^2$ denotes the squared Euclidean distance in three-dimensional space. $\sigma$ represents the standard deviation of the Gaussian function, which controls not only the width of the function but also serve as filter to filter out bounding box coordinates that are significantly distant from the box center predicted in the previous frame. 
After calculating the Gaussian mask of each point, in Equation~\eqref{gaussian mask operation on fusion features} it operating on features of last stage memory. The expression for the operation is as follows:
\begin{equation}
F = G_{m}^{n} \cdot (F_{G} + F_{M}).
\label{gaussian mask operation on fusion features}
\end{equation}
 Here, $F$ represents the features of applying the normalized Gaussian mask $G_{m}^{n}$ in a weighted manner to the fusion features of geometric features $F_{G}$ and mask features $F_{M}$, both of which are output from the memory of the bi-directional cross-frame memory module. The Gaussian mask as a distractor fiter to ensure precise and reliable tracking of points.

\textbf{Objectness Score Prediction Sampling.} We treat each target center as a proposal center and utilize features refined by a Gaussian mask to perform Hough Voting~\citep{Votenet}, thereby predicting the target center and its associated objectness score. Subsequently, we adopt a sampling strategy based on the predicted objectness score, which can focus on target center compared with farthest point sampling method. This process is described by the following equation:
\begin{equation}
C_{\text{p}} = \text{Sampling}(C, S^{n}_{topK}),
\end{equation}
where, $C_{\text{p}}$ denotes the candidate proposal center, $C$ and $S^{n}_{topK}$ represent the Hough Voting predicted object center and the normalized objectness score, respectively. We normalize the objectness score and rank these scores to select points of top $K$ objectness score, which can best represent the object. This sampling strategy not only yields more accurate proposals for precise tracking compared to farthest point sampling but also leverages pre-computed objectness scores, avoiding extra computational costs. 

\section{Experiments}
\subsection{Experimental Setup}
Dataset Selection. In this study, we utilized three widely recognized large-scale datasets: KITTI~\citep{KITTI}, NuScenes~\citep{NuScenes}, and Waymo Open Dataset~\citep{Waymon} (referred to as WOD), to demonstrate the practicality of our model. The KITTI dataset comprises 21 training video sequences and 29 testing video sequences. Due to the unavailability of test labels, in line with previous research~\citep{P2B}, we divided the training set into three parts: sequences 0-16 for training, 17-18 for validation, and 19-20 for testing. The NuScenes dataset presents greater challenges than KITTI, as it contains more data, with 700, 150, and 150 scenes for training, validation, and testing, respectively. For the WOD dataset, we categorized 1121 tracking sequences into easy, medium, and hard categories based on the density of point clouds, following the classification method of LiDAR-SOT~\citep{Lidar-sot}.
\subsection{Evaluation Criteria} 
Our study adheres to the One-Pass Evaluation method~\citep{One-Pass}. Success is measured based on the predicted and real bounding boxes, defined as the Area Under the Curve (AUC) of the plot displaying the proportion of frames where the Intersection Over Union (IOU) exceeds a certain threshold, ranging from 0 to 1. Precision is determined as the AUC of the plot showing the proportion of frames where the distance between the centers of the bounding boxes is within a specified threshold, ranging from 0 to 2 meters.

\subsection{Comparison with State-of-the-arts}
\noindent\textbf{Results on KITTI.} we present a comprehensive comparison of our method with the previous state-of-the-art approaches, namely SC3D~\citep{SC3D}, P2B~\citep{P2B}, PTT~\citep{PTT}, 3D-SiamRPN~\citep{3D-SiamRPN}, BAT~\citep{BAT}, LTTR\citep{LTTR}, V2B~\citep{V2B}, MLVSNet~\citep{MLVSNET}, C2FT~\citep{C2FT}, MLSET~\citep{MLSET},PTTR~\citep{PTTR}, Cui,et.al~\citep{Cui22}, TAT~\citep{TAT}, M2Track~\citep{M2Track}, CXTrack~\citep{CXTrack}, MBPTrack~\citep{MBPTrack} on the KITTI dataset. The published results from corresponding papers are reported. As depicted in Table~\ref{kitti}, our method surpasses previous state-of-the-art method MBPTrack of $0.3/0.3$ in average performance. We speculate that comprared with MBPTrack~\citep{MBPTrack} which depends on memory across two frames and box prior localizaiton netwrok. Our method can overcome the tracker drift better in pedestrain and cyclist. However, for large targets with more uniform shapes like van, the results show stability in success rates, with a notable increase in precision. This underlines the efficacy of our bi-directional memory mechanism, which, by compensating for missing information and augmenting data for unocclusioned targets, which helps locate target accurately. Despite this, our method still shows consistent performance gains across all categories, suggesting that our spatial-temporal tracking model is particularly advantageous for long-term tracking of small targets in complex scenes.

\begin{table}[t]
\renewcommand{\arraystretch}{1.0}
\renewcommand\tabcolsep{8.2pt} 
\footnotesize
\caption{\textbf{Comparisons with the state-of-the-art methods on KITTI dataset}. ``Mean'' is the average result weighted by frame numbers. ``\underline{Underline}'' and ``\textbf{Bold}'' denote previous and current best performance, respectively. Success/Precision are used for evaluation.}
\vspace{-0.3cm}
\begin{center}
\label{kitti}
\begin{tabular}{c|c|c|c|c|c}
\hline 
\multirow{2}{*}{Method} & Car & Pedestrian & Van & Cyclist & Mean\\
&(6424) & (6088) & (1248) & (308) & (14068) \\
\hline
 SC3D & 41.3/57.9 & 18.2/37.8 & 40.4/47.0 & 41.5/70.4 & 31.2/48.5 \\
 P2B & 56.2/72.8 & 28.7/49.6 & 40.8/48.4 & 32.1/44.7 & 42.4/60.0 \\ 
 3DSiamRPN & 58.2/76.2 & 35.2/56.2 & 45.7/52.9 & 36.2/49.0 & 46.7/64.9 \\
 LTTR & 65.0/77.1 & 33.2/56.8 & 35.8/45.6 & 66.2/89.9& 48.7/65.8 \\
 MLVSNet & 56.0/74.0 & 34.1/61.1 & 52.0/61.4 & 34.3/44.5 & 45.7/66.7 \\ 
 BAT & 60.5/77.7 & 42.1/70.1 & 52.4/67.0 & 33.7/45.4 & 51.2/72.8 \\
 PTT & 67.8/81.8 & 44.9/72.0 & 43.6/52.5 & 37.2/47.3 & 55.1/74.2 \\
 V2B & 70.5/81.3 & 48.3/73.5 & 50.1/58.0 & 40.8/49.7 & 58.4/75.2 \\
 CMT & 70.5/81.9 & 49.1/75.5 & 54.1/64.1 & 55.1/82.4 & 59.4/77.6 \\
 PTTR & 65.2/77.4 & 50.9/81.6 & 52.5/61.8 & 65.1/90.5 & 57.9/78.1 \\ 
 STNet & 72.1/84.0 & 49.9/77.2 & 58.0/70.6 & 73.5/93.7 & 61.3/80.1 \\
 TAT & 72.2/83.3 & 57.4/84.4 & 58.9/69.2 & 74.2/93.9 & 64.7/82.8\\ 
 M2-Track & 65.5/80.8 & 61.5/88.2 & 53.8/70.7 & 73.2/93.5 & 62.9/83.4 \\ 
 CXTrack & 69.1/81.6 & 67.0/91.5 & 60.0/71.8 & 74.2/94.3 & 67.5/85.3 \\
 MBPTrack & 73.4/84.8 & 68.6/93.9 & 61.3/72.7 & 76.7/94.3 & 70.3/87.9 \\
\hline
 STMD-Tracker (Ours) & \textbf{73.7}/\textbf{85.2} & \textbf{68.9}/\textbf{94.2} & \textbf{61.4}/72.7 & \textbf{76.9}/\textbf{94.9} & \textbf{70.6}/\textbf{88.2} \\
Improvement &\textcolor[rgb]{0.0,0.5,0.0}{\textit{$\uparrow$0.3}} /\textcolor[rgb]{0.0,0.5,0.0}{\textit{$\uparrow$0.4}} &\textcolor[rgb]{0.0,0.5,0.0}{\textit{$\uparrow$0.6}} /\textcolor[rgb]{0.0,0.5,0.0}{\textit{$\uparrow$0.3}} &\textcolor[rgb]{0.0,0.5,0.0}{\textit{$\uparrow$0.1}} /\textcolor[rgb]{0.0,0.5,0.0}{\textit{$\uparrow$0}} &\textcolor[rgb]{0.0,0.5,0.0}{\textit{$\uparrow$0.2}}/\textcolor[rgb]{0.0,0.5,0.0}{\textit{$\uparrow$0.6}} &\textcolor[rgb]{0.0,0.5,0.0}{\textit{$\uparrow$0.3}}/\textcolor[rgb]{0.0,0.5,0.0}{\textit{$\uparrow$0.3}}\\
\hline
\end{tabular} 
\end{center}
\vspace{-0.5cm}
\end{table}

\begin{figure*}[t]
\centering
\includegraphics[width=1\linewidth,height=5.0cm]{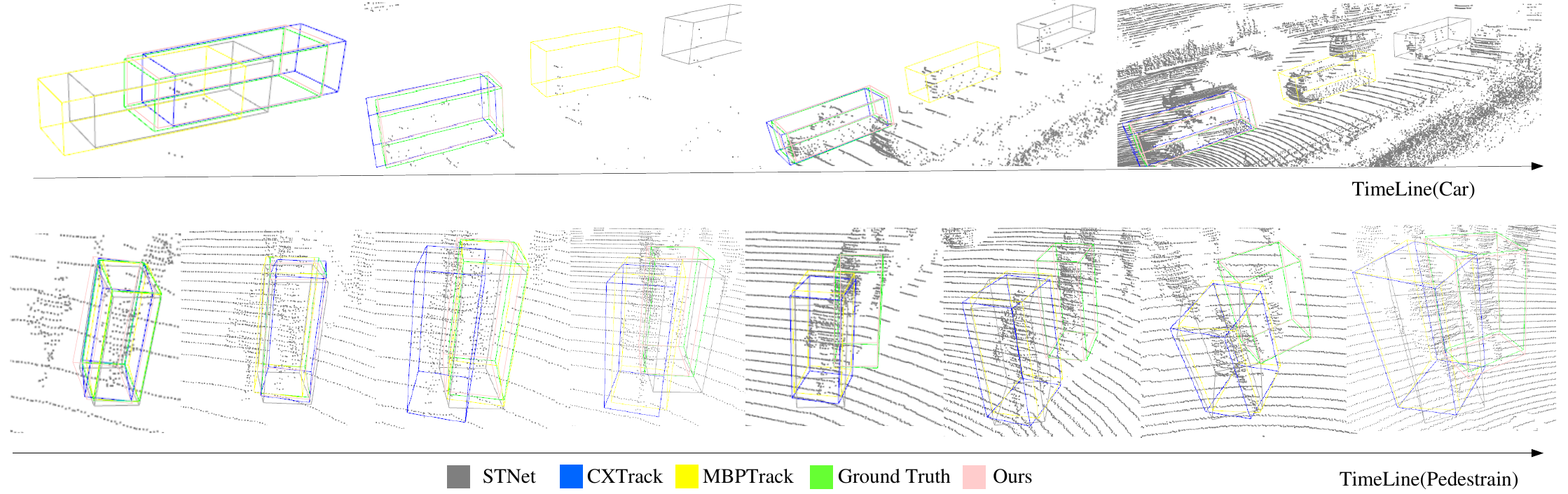}
\vspace{-0.7cm}
\caption{\textbf{Visualization of tracking results compared with state-of-the-art methods.}}
\label{fig:visualizaiton}
\vspace{-0.2cm}
\end{figure*}

For further explanation, we present a visual analysis of the tracking results on KITTI. As shown in Figure.~\ref{fig:visualizaiton}, MBPTrack~\citep{MBPTrack} and STNet~\citep{STNet} experience drift from the target car. In subsequent frames CXTrack~\citep{CXTrack} and ours can predict the orientation accurately on the Car category, while the predicted bounding boxes by our method hold tight to the ground truths. That's because our bi-directional memory mechanism can accurately adjust the tracking object's orientation. For pedestrians, all methods drift to nearby person due to the large appearance variation caused by heavy occlusion. However, only our method can track the target through intermittent occlusions and clearances, owing to the temporal convolution on multi frames, bi-directional cross-frame memory mechnasim and gaussian mask distractor filtering.

\noindent\textbf{Results on Waymon.} Moreover, we tested our method on the WOD dataset, employing models that had been pretrained on the KITTI dataset, following previous work. The corresponding categories between KITTI and WOD datasets are Car→Vehicle and Pedestrian→Pedestrian. The experimental results, as shown in Tab.~\ref{waymo}, illustrate that our approach achieves best tracking outcomes that surpass all other methods across various degrees of point sparsity. Particularly for pedestrians, our method surpasses all others in both precision and success rates across each category, regardless of their complexity, achieving an increase of over 1. For cars and pedestrians combined, there is an average improvement of 1.2 in success rates and 0.9 in precision. To summarize, the method we propose is capable of accurately identifying targets of any size and demonstrates robust performance in novel, unseen scenarios. This further validates the effectiveness of our approach, where the spatio-temporal convolutional backbone not only represents spatial contextual features but also captures temporal variations. The bi-directional memory mechanism compensates for information lost due to appearance variance from occlusion, and the Gaussian mask filters out distractor to improve tracking performance.
\begin{table*}[t]
\renewcommand{\arraystretch}{1.0}
\renewcommand\tabcolsep{8.2pt} 
\footnotesize
\begin{center}
\caption{\textbf{Comparison with state of the arts on Waymo Open Dataset.}}
\vspace{-0.2cm}
\label{waymo}
\resizebox{1\linewidth}{!}{
\begin{tabular}{c|cccc|cccc|c}
\hline 
\multirow{2}{*}{Method} & \multicolumn{4}{c|}{Vehicle(185731)} & \multicolumn{4}{c|}{Pedestrian(241752)} & \multirow{2}{*}{Mean(427483)}\\
& Easy & Medium & Hard & Mean & Easy & Medium & Hard & Mean & \\
\hline
 P2B& 57.1/65.4 & 52.0/60.7 & 47.9/58.5 & 52.6/61.7 & 18.1/30.8 & 17.8/30.0 & 17.7/29.3 & 17.9/30.1 & 33.0/43.8\\ 
 BAT& 61.0/68.3 & 53.3/60.9 & 48.9/57.8 & 54.7/62.7 & 19.3/32.6 & 17.8/29.8 & 17.2/28.3 & 18.2/30.3 & 34.1/44.4\\
 V2B& 64.5/71.5 & 55.1/63.2 & 52.0/62.0 & 57.6/65.9 & 27.9/43.9 & 22.5/36.2 & 20.1/33.1 & 23.7/37.9 & 38.4/50.1\\
 STNet& 65.9/72.7 & 57.5/66.0 & 54.6/64.7 & 59.7/68.0 & 29.2/45.3 & 24.7/38.2 & 22.2/35.8 & 25.5/39.9 & 40.4/52.1\\ 
 TAT& 66.0/72.6 & 56.6/64.2 & 52.9/62.5 & 58.9/66.7 & 32.1/49.5 & 25.6/40.3 & 21.8/35.9 & 26.7/42.2 & 40.7/52.8\\
 CXTrack & 63.9/71.1 & 54.2/62.7 & 52.1/63.7 & 57.1/66.1 & 35.4/55.3 & 29.7/47.9 & 26.3/44.4 & 30.7/49.4 & 42.2/56.7\\
 M2Track & 68.1/75.3 & 58.6/66.6 & 55.4/64.9 & 61.1/69.3 & 35.5/54.2 & 30.7/48.4 & 29.3/45.9 & 32.0/49.7 & 44.6/58.2\\
MBPTrack & 68.5/77.1 & 58.4/68.1 & 57.6/69.7 & 61.9/71.9 & 37.5/57.0 & 33.0/51.9 & 30.0/48.8 & 33.7/52.7 & 46.0/61.0\\
\hline
STMD-Tracker (Ours) & \textbf{69.4}/\textbf{78.2} &\textbf{58.7}/\textbf{68.5} & \textbf{58.3}/\textbf{70.6} & \textbf{62.1}/\textbf{72.4} & \textbf{38.7}/\textbf{59.1} & \textbf{35.2}/\textbf{53.3} & \textbf{31.8}/\textbf{50.4} & \textbf{35.2}/\textbf{54.3} & \textbf{46.9}/\textbf{62.2}\\
Improvement &\textcolor[rgb]{0.0,0.5,0.0}{\textit{$\uparrow$0.9}} /\textcolor[rgb]{0.0,0.5,0.0}{\textit{$\uparrow$1.1}} 
&\textcolor[rgb]{0.0,0.5,0.0}{\textit{$\uparrow$0.3}} /\textcolor[rgb]{0.0,0.5,0.0}{\textit{$\uparrow$0.4}} 
&\textcolor[rgb]{0.0,0.5,0.0}{\textit{$\uparrow$0.7}} /\textcolor[rgb]{0.0,0.5,0.0}{\textit{$\uparrow$0.9}} 
&\textcolor[rgb]{0.0,0.5,0.0}{\textit{$\uparrow$0.2}}/\textcolor[rgb]{0.0,0.5,0.0}{\textit{$\uparrow$0.5}} 
&\textcolor[rgb]{0.0,0.5,0.0}{\textit{$\uparrow$1.2}}/\textcolor[rgb]{0.0,0.5,0.0}{\textit{$\uparrow$2.1}}
&\textcolor[rgb]{0.0,0.5,0.0}{\textit{$\uparrow$2.2}}/\textcolor[rgb]{0.0,0.5,0.0}{\textit{$\uparrow$1.4}}
&\textcolor[rgb]{0.0,0.5,0.0}{\textit{$\uparrow$1.8}}/\textcolor[rgb]{0.0,0.5,0.0}{\textit{$\uparrow$1.6}}
&\textcolor[rgb]{0.0,0.5,0.0}{\textit{$\uparrow$1.5}}/\textcolor[rgb]{0.0,0.5,0.0}{\textit{$\uparrow$1.6}}
&\textcolor[rgb]{0.0,0.5,0.0}{\textit{$\uparrow$0.9}}/\textcolor[rgb]{0.0,0.5,0.0}{\textit{$\uparrow$1.2}}\\
\hline
\end{tabular} }
\end{center}
\vspace{-0.3cm}
\end{table*}

\noindent\textbf{Results on Nuscenes.}
As illustrated in Tab.~\ref{nuscene}, our STMD-Tracker obtains the best performance in nuScenes dataset for all categories and demonstrates superior results compared with other methods~\citep{SC3D,P2B,BAT,Cui22,MLSET}. The multi-frames method~\citep{MBPTrack} will tend to drift intra-class distractor. To address this issue, we designed a bi-directional cross-frame memory by incorporating future frames to compensate for the current loss of information. Unlike other methods that focus on learning target and distractor feature representations or being aware of distractors, our method employs a Gaussian mask prior to filter out distractor points in 3D coordinates. We think alought motion-based M2Track~\citep{M2Track} is better than mathching-based method, the inaccurate segemtation and coordinates transformation will limit tracking performance. In cotrast, our method leverage Gaussian mask prior to help locate the target accurately. Notably, our STMD-Tracker outperforms MBPTrack~\citep{MBPTrack} in the Pedestrian category by $1.54$ in Success and $1.24$ in Precision. Our method also achieves notable improvements for bus and cars because our tracker utilizes spatial and temporal convolution to model multi-frame relationships and bi-directional memory to augment target points. Our method leverages a spatial-temporal graph backbone complemented by cross-frame memory and a Gaussian mask to filter out distractor, which seems to be particularly effective for smaller targets with complex geometries. 
\begin{table*}[t]
\begin{center}
\caption{\textbf{Comparisons with the state-of-the-art methods on NuScenes dataset}. }
\vspace{-0.3cm}
\label{nuscene}
\resizebox{1.0\linewidth}{!}{
\begin{tabular}{c|c|c|c|c|c|c}
\hline 
Method & Car(64159) & Pedestrian(33227) & Truck(13587) & Trailer(3352) & Bus(2953) & Mean(117278) \\
\hline
 SC3D & 22.31/21.93 & 11.29/12.65 & 30.67/27.73 & 35.28/28.12 & 29.35/24.08 & 20.70/20.20 \\
 P2B & 38.81/43.18 & 28.39/52.24 & 42.95/41.59 & 48.96/40.05 & 32.95/27.41 & 36.48/45.08 \\ 
MLSET & 53.20/58.30 & 33.20/58.60 & 57.36/52.50 & 57.61/40.90 & / & / \\
 SMAT & 43.51/49.04 & 32.27/60.28 & 44.78/44.69 & 37.45/34.10 & 39.42/34.32 & 40.20/ 50.92 \\
BAT & 40.73/43.29 & 28.83/53.32 & 45.34/42.58 & 52.59/44.89 & 35.44/28.01 & 38.10/45.71 \\
 M2-Track & 55.85/65.09 & 32.10/60.92 & 57.36/59.54 & 57.61/58.26 & 51.39/51.44 & 49.23/62.73 \\ 
 MBPTrack & 62.47/70.41 & 45.32/74.03 & 62.18/63.31 & 65.14/61.33 & 55.41/51.76 & 57.48/69.88\\
  \hline
  STMD-Tracker (Ours) & \textbf{63.05}/\textbf{71.24} & \textbf{46.86}/\textbf{75.27} & \textbf{62.87}/\textbf{63.96} & \textbf{65.24}/\textbf{62.03} & \textbf{56.02}/\textbf{52.88} & \textbf{58.33}/\textbf{70.81}\\
  Improvement &\textcolor[rgb]{0.0,0.5,0.0}{\textit{$\uparrow$0.58}} /\textcolor[rgb]{0.0,0.5,0.0}{\textit{$\uparrow$0.83}} 
&\textcolor[rgb]{0.0,0.5,0.0}{\textit{$\uparrow$1.54}} /\textcolor[rgb]{0.0,0.5,0.0}{\textit{$\uparrow$1.24}} 
&\textcolor[rgb]{0.0,0.5,0.0}{\textit{$\uparrow$0.69}} /\textcolor[rgb]{0.0,0.5,0.0}{\textit{$\uparrow$0.65}} 
&\textcolor[rgb]{0.0,0.5,0.0}{\textit{$\uparrow$0.1}}/\textcolor[rgb]{0.0,0.5,0.0}{\textit{$\uparrow$0.65}} 
&\textcolor[rgb]{0.0,0.5,0.0}{\textit{$\uparrow$0.61}}/\textcolor[rgb]{0.0,0.5,0.0}{\textit{$\uparrow$1.12}}
&\textcolor[rgb]{0.0,0.5,0.0}{\textit{$\uparrow$0.85}}/\textcolor[rgb]{0.0,0.5,0.0}{\textit{$\uparrow$0.93}}\\
\hline
\end{tabular} }
\end{center}
\vspace{-0.3cm}
\end{table*}

\subsection{Ablation Studies}
Based on KITTI datasets, we conduct comprehensive experiments in Car and Pedestrain categories to validate the effectiveness of our proposed method. 

\noindent\textbf{Model components.}
The $1^{st}$ row in Tab.~\ref{tab:ablation study} shows that the success/precision of MBPTrack\citep{MBPTrack} in car and pedestrain categories. The $2^{nd}$ row shows the application of temporal convolution (TC) for exploring multi-frames interactions in subsequent stage yields positive outcomes, especially for pedestrain. Furthermore, as indicated in the $3^{nd}$ row, the \textit{Mean} performance saw a significant improvement. Our bi-directional Cross-frame Memory (BCM) substantially compensates for the information loss due to rapid frame variation, a scenario where minimal variations between frames and their quick succession might prevent the module from fully capturing key information. Specifically for the car category, the unoccluded target point will be augmented due to the bi-directional memory mechanism. Subsequently, as demonstrated in the $4^{th}$ row, the incorporation of a Gaussian-mask (GM) prior alongside objectness score-based proposal sampling leads to an imporvement in \textit{Mean} performance by 0.32/0.39 demonstrating the effectiveness of our GM in enhancing the network's capability. 

\begin{table}[t]
\renewcommand\tabcolsep{4pt}
\begin{center}
\caption{\textbf{Ablation studies on model components.} TC means temporal convolution. BCM represents Bi-directional Cross-Frame Memory Module. GMD is Gaussian-Mask Distractor filtering Module.}~\label{tab:ablation study}
\vspace{-0.1cm}
\begin{tabular}{c|c|c|c|c|c}
\specialrule{.05cm}{0pt}{0pt}
TC & BCM & GM & Car & Pedestrian & Mean \\ \hline \hline
\texttimes  &\texttimes &\texttimes& 73.40 / 84.80 & 68.60 / 93.90 &  71.06 / 89.23 \\
\checkmark &\texttimes& \texttimes  & 73.50 / 84.94 & 68.77 / 94.01 & 71.20 / 89.35 \\
\checkmark & \checkmark &\texttimes  & 73.67 / 85.08 & 68.84 / 94.15 &  71.32 / 89.49 \\
\checkmark   & \checkmark   & \checkmark   & \textbf{73.72} / \textbf{85.24}  & \textbf{68.91}/ \textbf{94.25} & \textbf{71.38} / \textbf{89.62} \\ 
\specialrule{.05cm}{0pt}{0pt} 
\end{tabular} 
\end{center}
\vspace{-0.6cm}
\end{table}
\noindent\textbf{Temporal Convolution.}
The ablation study results presented in Table~\ref{tab:Temporal Convolution} highlight that performing convolution across 8 frames using the replicate padding strategy significantly outperforms the no padding and zero padding strategies, underscoring the crucial role of padding strategies in enhancing tracking outcomes. Specifically, replicate padding, by duplicating frames at the beginning and end of a sequence, not only enriches the convolutional context but is particularly effective for sequences with no more than 10 frames. This temporal convolution effectively preserves the integrity of target features, avoiding the dilution of target information that can occur with zero padding, and provides a significant advantage in achieving more accurate feature representation. Furthermore, the experimental data reveal the specific impact of padding strategies on performance across different frame counts, especially highlighting the advantages of replicate padding in scenarios with fewer frames. Compared to zero padding, which can offer some performance improvements in certain situations, replicate padding demonstrates superior performance in maintaining the integrity of target features and optimizing feature representation.

\begin{table}[!h]
\renewcommand{\arraystretch}{1}
\renewcommand\tabcolsep{2.0pt}
  \centering
  \footnotesize
  \caption{\textbf{Ablation study on padding mode between different frames.}}
  \begin{tabular}{c|cc|cc|cc}
  \specialrule{.05cm}{0pt}{0pt} 
  \multirow{2}{*}{Padding} & \multicolumn{2}{c|}{$8\_frames$} & \multicolumn{2}{c|}{$9\_frames$} & \multicolumn{2}{c}{$10\_frames$} \\
  &Car& Pedestrain&Car&Pedestrain&Car&Pedestrain\\
  \hline \hline \rule{0pt}{8pt}
 No  & / & / & / & / &73.2,84.7& 68.4,93.5\\
Zero &68.4,78.6 &66.4,91.1 &66.0,75.4  &62.8,88.5& / &/ \\
Replicate & \textbf{73.7,85.2} &\textbf{68.9,94.2} &67.2,77.5 &64.8,90.3& / & /\\
  \specialrule{.05cm}{0pt}{0pt} 
  \end{tabular}
  \label{tab:Temporal Convolution}
\end{table}

\noindent\textbf{Gaussian Mask Feature Fusion Ablation.}
Based on the results presented in Table~\ref{tab:gaussian mask}, where F1 represents geometric features derived from DeFPM~\citep{MBPTrack}, F2 embodies mask features, and F3 is the aggregated value of F1 and F2. It can be observed that applying the Gaussian mask prior to combined features F3 not only enhances the model's tracking success rate and precision for both car and pedestrian targets but also significantly surpasses the performance achieved by processing each feature individually. This substantial improvement can be attributed to the Gaussian mask's ability to more effectively capture the intrinsic correlations between geometric and mask features when processed together, providing the model with richer and more interconnected information. Additionally, our method underscores the importance of considering the potential interactions between different features in feature fusion strategies, indicating that integrating a variety of features can significantly optimize tracking performance and offers important guidance for future model design.

\begin{table}[!h]
\renewcommand{\arraystretch}{1.0}
\renewcommand\tabcolsep{8.0pt}
  \centering
  \footnotesize
  \caption{\textbf{Ablation studies of gaussian mask operation.}}
  \begin{tabular}{c|cc|cc}
  \specialrule{.05cm}{0pt}{0pt} 
  \multirow{2}{*}{Operation} & \multicolumn{2}{c|}{Success} & \multicolumn{2}{c}{Precision} \\
  & Car & Pedestrain & Car & Pedestrain \\
\hline \hline \rule{0pt}{8pt}
F1 & 72.1 & 83.9 & 68.1 & 93.7 \\
F2 & 71.5 & 82.6 & 67.0 & 91.2 \\
F3 & \textbf{73.7} & \textbf{85.2} & \textbf{68.9} & \textbf{94.2} \\
  \specialrule{.05cm}{0pt}{0pt} 
  \end{tabular}
  \label{tab:gaussian mask}
\end{table}
\noindent\textbf{Optimal Sigma Value for Gaussian Mask.}
As shown in Table~\ref{tab:sigma} the optimal performance not only achieved with a sigma value of 2 for the Gaussian mask but also elucidates the nuanced balance between specificity and generality in tracking applications. By selecting a sigma of 2, the Gaussian mask adeptly retains an optimal number of coordinate points, which is crucial for ensuring that the tracking algorithm focuses on the most relevant spatial data. Our method can effectively filters out irrelevant coordinates, thereby reducing the likelihood of the algorithm being misled by distractor that are spatially proximate but irrelevant to the tracking target. Moreover, our method significantly enhances the model's ability to distinguish between the target and similar distractor within the coordinate space, thereby improving both the success rate and precision for tracking cars and pedestrians. The results highlight the importance of fine-tuning the Gaussian mask's sigma value to achieve a delicate balance between capturing essential details and eliminating noise, which is pivotal for optimizing tracking accuracy in complex environments.
\begin{table}[t]
\renewcommand{\arraystretch}{1.0}
\renewcommand\tabcolsep{8.0pt}
  \centering
  \footnotesize
  \caption{\textbf{Ablation studies of sigma of gaussian mask.}}
  \begin{tabular}{c|cc|cc}
  \specialrule{.05cm}{0pt}{0pt}
  \multirow{2}{*}{Sigma} & \multicolumn{2}{c|}{Success} & \multicolumn{2}{c}{Precision} \\
  & Car & Pedestrain & Car & Pedestrain \\
\hline \hline \rule{0pt}{8pt}
0.5 & 68.5 & 77.6 & 58.6 & 82.3 \\
1.0 & 71.3 & 80.4 & 67.2 & 92.1 \\
1.5 & 72.1 & 83.5 & 68.4 & 93.0 \\
2.0 & \textbf{73.7} & \textbf{85.2} & \textbf{68.9} & \textbf{94.2} \\
  \specialrule{.05cm}{0pt}{0pt}
  \end{tabular}
  \label{tab:sigma}
  \vspace{-0.5cm}
\end{table}

\section{Conclusion}
We propose a spatio-temporal bi-directional cross-frame distractor filtering tracker, named STMD-Tracker, to address the issue of tracker drift due occlusion or distractor in 3D single object tracking. The STMD-Tracker employs a 4D decoupled spatial and temporal convolution backbone to exploit spatial contextual information and rich temporal variation in historical frames. Furthermore, we designed a bi-directional cross-frame memory process to compensate the loss of target information due to intra-class distractor or occlusion, the proposed memory also can augment correlation feature in unoccluded scenio. Finally, we designed a novel spatial reliable Gaussian mask localization network, named GMLocNet, which leverages Gaussian mask to filter out similar distractor and more accurately localize targets of various sizes. Extensive experiments on three large-scale datasets demonstrate that our method surpasses previous state-of-the-art methods in tracking targets of varying sizes. The major limitation of our work lies in the reduced success of tracking targets under extremely sparse conditions and at great distances. Additionally, during long-term tracking, repeated instances of tracker drift lead to the tracker's ultimate inability to track the target effectively. In the future, we aim to explicitly model the target's long-range dependencies to better address these issues.

\section*{Acknowledgements}
We are grateful for the financial support provided by \emph{National Key R\&D Program of China} and \emph{Xi’an Key Laboratory of Active Photoelectric Imaging Detection Technology}.
\section*{Disclosure statement}
No potential conflict of interest was reported by the author(s).
\section*{Funding}
This research was funded by \emph{National Key R\&D Program of China}, under grant number \emph{2022YFC3803702}.

\bibliographystyle{chicagoa}
\bibliography{reference} 

\end{document}